\documentclass[conference,compsoc]{IEEEtran}
\usepackage[utf8]{inputenc}
\usepackage[T1]{fontenc}
\IEEEoverridecommandlockouts
\usepackage{amsmath,amssymb,amsfonts}
\usepackage{graphicx}
\usepackage{textcomp}
\usepackage{xcolor}
\usepackage{hyperref}

\newcommand{\code}{\texttt}

\usepackage{xspace}
\usepackage{todonotes}

\usepackage{subcaption}

\usepackage{booktabs}
\usepackage{multirow}

\usepackage[ruled]{algorithm2e}

\usepackage[backend=biber,style=alphabetic]{biblatex}
\newcommand{\ownset}[1]{\ensuremath{\mathcal{#1}}\xspace}
\newcommand{\AR}{\ownset{A}}
\newcommand{\MR}{\ownset{M}}
\newcommand{\WR}{\ownset{W}}
\newcommand{\SR}{\ownset{S}}
\newcommand{\IR}{\ownset{I}}
\newcommand{\allfeatures}{\ownset{G}}

\newcommand{\fmes}{\ensuremath{F_1}\xspace}

\newcommand{\data}{X}
\newcommand{\datum}{x}
\newcommand{\datumi}{i}
\newcommand{\target}{y}
\newcommand{\feat}{\ensuremath{j}}

\newcommand{\numsamples}{n}
\newcommand{\numfeatures}{d}

\newcommand{\excepti}[1][i]{\allfeatures \setminus #1}
\newcommand{\dataexi}[1][i]{\data_{\excepti[#1]}}
\newcommand{\func}{f}
\newcommand{\otherfunc}{h}
\newcommand{\optfunc}{\bar{\func}}
\newcommand{\allfuncs}{\mathcal{F}}
\newcommand{\allfuncssamescore}{\bar{\mathcal{F}}}
\newcommand{\realfunc}{g}

\newcommand{\lossname}{\mathcal{L}}
\newcommand{\lossfunc}[2]{\lossname(#1,#2)}

\newcommand{\ie}{,\ i.e.\ }

\newcommand{\permname}{p}
\newcommand{\perm}[1]{\permname_{#1}(\data)}

\newcommand{\datawithperm}{\data^\star}
\newcommand{\numscoresamples}{\alpha}

\newcommand{\losssamplesname}{\widehat{\pi}}
\newcommand{\probedist}[1]{\losssamplesname(#1)}
\newcommand{\predi}{\Pi}
\newcommand{\PI}[1]{\predi(#1, \numscoresamples):= \displaystyle {\overline {\probedist{\cdot}}}_{\numscoresamples}\pm T_{\numscoresamples-1}(p) \cdot \sigma(\probedist{\cdot}) \sqrt {1+(1/\numscoresamples)}}

\DeclareMathOperator{\imp}{imp}
\newcommand{\featimpfunc}[2]{\imp(#1,#2)}
\newcommand{\impsamplesname}{\widehat{\gamma}}

\newcommand{\shadowfeatprobedist}[1]{\widehat{\gamma_s}(#1)}
\newcommand{\shadowfeatpredi}{\Gamma_s}
\newcommand{\shadowPI}[1]{\shadowfeatpredi(#1, \numscoresamples):= \displaystyle {\overline {\shadowfeatprobedist{\cdot}}}_{\numscoresamples}\pm T_{\numscoresamples-1}(p) \cdot \sigma(\shadowfeatprobedist{\cdot}) \sqrt {1+(1/\numscoresamples)}}

\begin{document}

\title{Sequential Feature Classification in the Context of Redundancies\\
    \thanks{This contribution has been made possible through
        the Bielefeld Young Researchers' Fund and the BMBF
        (grant number 01S18041A)}
}
\author{\IEEEauthorblockN{Lukas Pfannschmidt\IEEEauthorrefmark{1},
        Barbara Hammer\IEEEauthorrefmark{2}}
    \IEEEauthorblockA{\textit{Machine learning group} \\
        \textit{Bielefeld University}\\
        Bielefeld, Germany\\
        \IEEEauthorrefmark{1}lpfannschmidt@techfak.uni-bielefeld.de,
        \IEEEauthorrefmark{2}bhammer@techfak.uni-bielefeld.de}}

\maketitle

\begin{abstract}
The problem of all-relevant feature selection is concerned with finding a relevant feature set with preserved redundancies.
There exist several approximations to solve this problem but only one could give distinction between strong and weak relevance.
This approach was limited to the case of linear problems.
In this work we present a new solution for this distinction in the non-linear case through the use of random forest models and statistical methods.
\end{abstract}

\begin{IEEEkeywords}
    feature selection, interpretability, redundancies, strong and weak relevance
\end{IEEEkeywords}

\section{Background}
\label{sec:background}

To interpret the behaviour of machine learning models it is important to find the original input features which correspond to the output.
Often the majority of input features is irrelevant, and we seek a feature subset which only contains relevant features in relation to an unknown underlying process.
Furthermore, a feature subset without redundancies can be beneficial by increasing robustness.
Such a subset of features with minimal redundant information is called the minimal feature subset and the majority of existing feature selection methods belong to the minimal feature selection problem.

In many cases, the function or relation between input features is unknown.
This is often encountered in the life sciences where machine learning is used on data sets without identifying all variables beforehand.
Here, the main goal of machine learning shifts from making perfect predictions to identifying the contributing elements.
This is in part contrary to the minimal feature selection problem which removes redundant features in the process.
These features could lead to additional insights in the biological mechanisms.

The \emph{all-relevant feature selection problem} (ARFS) is the task of finding a relevant feature subset including redundant features.
To find the all-relevant feature set, denoted as \AR, is often much harder than finding the minimal relevant set \MR.
To identify \MR many approaches exist, such as the Lasso \autocite{tibshirani_regression_1996} and ElasticNet \autocite{elasticnet}, which exploit regularization to achieve sparseness in the model's use of features and thus produce a compressed subset.
In the presence of redundancies these subset and model solutions only represent one of many possible solutions and are not unique.
To find \AR one would have to find all possible models, which is infeasible in practise for the exact solution.

Consequently, several approximations exist and where compared in \autocite{degenhardtEvaluationVariableSelection}.
One of the best performing approaches is the Boruta wrapper method \autocite{kursaFeatureSelectionBoruta2010}.
It utilizes a random forest model together with random \emph{shadow} features, which are used as comparison variables, to filter out all irrelevant features while keeping redundant features.
Another new approach not compared in \autocite{degenhardtEvaluationVariableSelection} utilized feature relevance bounds to get an approximation of all possible feature contributions and then utilized random contrast features for approximation of a relevance threshold \autocite{gopfert2018}.
Not only were these bounds able to produce an approximation of \AR but could also classify for each relevant feature if other redundant features existed (weak relevance) or if they were unique in their information content (strongly relevant).
This made it possible to get insights into possible relations between weakly relevant features \autocite{pfannschmidtFRIFeatureRelevanceIntervals2019}.
The feature relevance bounds could be phrased in terms of efficient convex linear problems and were released as a standalone analysis tool named FRI \autocite{pfannschmidtLpfannFriFeature2020}.
While the linear problems are efficient the approach does not scale to bigger data sets and can not be adapted to non-linear problems without losing convexity of the optimization.

In this work we present a method which extends the Boruta method with a feature classification step which produces the same discrimination between strong and weak relevance as seen in \autocite{pfannschmidtFRIFeatureRelevanceIntervals2019}.
We therefore combine the advantages of Boruta and FRI while achieving more efficient run times.
Relevance decisions can be made based on the accuracy of the model when excluding a single feature in question.
In \autoref{sec:rfs} we describe our approach of efficiently decomposing the overall feature set into subsets with different characteristics to produce the distinction between strong and weak relevant features without testing all available features.
We improve existing methods to find set \MR by using statistical based thresholds and introduce a robust score testing scheme in \autoref{sub:loss_comparison} to improve feature classification.
In \autoref{sec:results} we analyse the general feature selection performance against other established approaches in context of redundant features.

\newpage

\section{Methods}
\label{sec:methods}

\subsection{Redundant Feature Selection}
\label{sec:rfs}
Let $\data$ be data set $\data:=\{\datum_{\datumi} \in \mathbb{R}^\numfeatures; \datumi=1 \dots \numsamples \}$ with $\numsamples$ samples and with $\allfeatures := \{j \in \mathbb{Z}; \feat=1 \dots \numfeatures\}$ as the set of all features such that cardinality $|\allfeatures| = \numfeatures$.
Target variable $\target \in \mathbb{R}^\numsamples$ is distributed according to some unknown function dependent on $\data$ such that $\realfunc(\data)=\target$.
Without loss of generality we limit $\target$ to be continuous and $\realfunc$ to be a regression function. 
Let $\optfunc := \hat \realfunc$ be the optimal estimator of $\realfunc$ with minimal estimation loss.
Consider the estimation loss as $$\lossfunc{\data}{\func}:= \sum_{\forall \datumi} \left| \func(\datum_{\datumi})- \target_\datumi \right|$$
where $\datum \in \data$ which denotes the deviation from $\realfunc$ if no random noise is involved.

We are interested in the input features from $\data$ which are relevant to $\func$ as defined by \textcite{nilssonConsistentFeatureSelection}.
One can define the relevance of a feature $\feat$ for a single model such as $\optfunc$ or for all models in a given set e.g. $\func \in \allfuncs$ as the set of all possible models.
In the context of redundancies we are interested in all functions with similar $\lossname$ and possibly different composition of input features:
$$\allfuncssamescore := \{ \func \in \allfuncs \ |\  \lossfunc{\data}{\func} \approx \lossfunc{\data}{\optfunc}  \}$$
Before we observed the set of data $\data$ with all features in $\allfeatures$\ie $\data = \data_\allfeatures$,
now we also consider the data set with specific subsets of features, e.g. the dataset with feature $\feat$ removed is denoted as $\dataexi[\feat]$

\Textcite{kohavi_wrappers_1997} introduced several classes of feature relevance and the resulting subsets which the general feature set is composed of.
These are the set of strongly relevant features (\SR), the set of weakly relevant features (\WR) and irrelevant features (\IR).
The union of \SR and \WR is the set of all-relevant features \AR:  $$\AR := \SR \cup \WR.$$
The goal of all relevant feature selection is finding all features belonging to \AR, and not to identify the detailed composition of \SR and \WR.
Most existing methods only produce information about membership of \AR, except the feature relevance bounds method \autocite{gopfert2018}, which yields both \SR and \WR in the linear case.

One existing all-relevant selection method is called Boruta \autocite{kursaFeatureSelectionBoruta2010}.
It builds on the information metrics acquired by observing the single trees of a random forest model.
Because the scores are not consistent with their real significance \autocite{rudnickiStatisticalMethodDetermining2006}, Boruta employs an extended information system.
Extended in the way, that additional to normal features Boruta adds shadow features.
Shadow features are randomly shuffled clones of existing features to remove correlation with the target.
Through the addition of those, one can estimate a contrast distribution of features, which are by design irrelevant.
It then tests the real features against the shadow features iteratively, increasing the significance threshold, until all features are tested conclusively.
Boruta's comparison with a null distribution inspired one aspect of our proposed method, which we come back to later.
More importantly, by comparing with a null distribution multiple times and introducing stochastic noise by repeatedly running a random forest model, Boruta can identify the \AR set with high precision.

Having found \AR, one can trivially classify all features in $\allfeatures$ which are not in \AR, as irrelevant:
$$\IR := \allfeatures \setminus \AR  $$
To decompose \AR into \SR and \WR we have to identify membership with at least one of them.
We can check for strong relevance by repeatedly fitting models and checking the behaviour of the loss function similar to the approach in \autocite{nilssonConsistentFeatureSelection}.
A feature $\feat$ is strongly relevant if  $$\min_{\func} \lossfunc{\dataexi[\feat]}{\func} > \min_{\otherfunc} \lossfunc{\data}{\otherfunc}.$$
This comparison would have to be performed for all $\feat \in \allfeatures$.
While $|\AR| \ll |\allfeatures|$ in most cases\ie the feature space is often sparse and most of the features can be ignored, we still can improve on this naive approach.

Earlier, we considered the specific problem of all-relevant feature selection.
In general when feature selection is performed, one considers the \emph{minimal-optimal} feature set (\MR).
The problem of finding \MR can be stated as finding the minimal non-redundant feature subset with the maximum amount of information for some function class.
This type of feature selection is easier than the ARFS~\cite{nilssonConsistentFeatureSelection} and many efficient methods exist such as greedy recursive feature elimination (RFE) \autocite{gregoruttiCorrelationVariableImportance2017}.

In the following we are using a new efficient threshold based approach which decides based on the models' importance values which features are important.
Importance values are internal parameters which correspond to input features similar to the weights of linear models and schemes such as Lasso.
Because we utilize Borute, we consider the importance values of a random forest model.
In \autoref{sub:importance_statistics} we describe in detail how we find \MR  with a sparse parameterization and statistical method to overcome inconsistent importance values.
For now, we consider \MR as given by this efficient approach, but alternatives such as RFE could also be applied.

By definition, $\MR \subset \AR$ and $\SR \subset \MR$.
All features in the minimal optimal set \MR are also included in \AR and furthermore  all strongly relevant features are included in \MR.
Instead of iterating and testing all features in $\AR$ we only have to consider the features in \MR.
In most cases when redundant relevant features are preset\ie $|\MR| \ll |\AR|$, it is much more efficient to only check features in the subset \MR.

Therefore, it is sufficient to identify \SR in \MR through comparing the loss after the elimination of each feature and identify \WR through
\begin{equation}\label{eq:decomposition}
    \WR := \AR \setminus \SR.
\end{equation}
Having said that, comparison of the loss is not straightforward as it requires a robust threshold to test against.

\subsection{Robust Loss Comparison}
\label{sub:loss_comparison}
In practise, we cannot perform a simple comparison between a reduced (set with feature $\feat$ removed) and a normal feature set.
Due to the stochastic nature of random forests, even fitting the same data set without feature removal can lead to variable models and thus to differing average losses.
To counteract this, we estimate a distribution of $\lossname$.
The distribution should represent a range of possible values which we regard as insignificant changes, similar to how our model would change, if an irrelevant feature would be removed.

Remember that we fit a RF model on a reduced dataset where one feature is eliminated and then observe the loss of the resulting model.
The distribution should emulate this given setting and as such we would have to eliminate one feature in our sampling procedure as well.
However, we cannot remove any feature from the initial feature set or we could possibly remove another relevant feature by chance.
To emulate the changes in model size we therefore permute a randomly chosen feature and add it to the dataset.
Permutation of feature $\feat$ is denoted as $\perm{\feat}$ and an extended dataset with such a permuted feature as $$\datawithperm := \{\data \cup {\perm{\feat} |\; \feat \sim U(\numfeatures)} \}.$$
$\perm{\feat}$ has no dependence on the target variable and represents an irrelevant feature.

We then define the random population
\[
    \probedist{\optfunc,\data}:=\{\lossfunc{\datawithperm}{\optfunc}\}^{\numscoresamples}
\] with parameter $\numscoresamples \in \mathbb{Z}$ as the number of samples used.

We then define an interval of plausible values based on a t-distribution 
\[
    \PI{\cdot}.
\]
Here $\overline {\pi}_{\numscoresamples}$ denotes the sample mean and $\sigma(\cdot)$ the standard deviation, and $T$ represents Student's t-distribution with $\numscoresamples-1$ degrees of freedom.
The size of $\predi$ depends on parameter $p \ll 1$ and the number of samples.
In our experiments $\numscoresamples\geq50$ already yields robust thresholds for common feature set sizes.
The interval only has to be computed once and is valid for all feature comparisons as it represents the distribution of irrelevant features and not a feature specific one. 

Feature $\feat$ is strongly relevant if $$\lossfunc{\dataexi[\feat]}{\optfunc} < \predi(\optfunc,\data)$$ which means we only have to check the lower bound of the prediction interval.

Using this procedure and checking all features in \MR leads to the complete set of strongly relevant features \SR and therefore also \WR as seen in \autoref{eq:decomposition}.

\begin{algorithm}
    \DontPrintSemicolon
    \SetKwData{scores}{$\losssamplesname$}
    \SetKwData{samplescore}{loss}
    \SetKwData{allsampleirrelimps}{$\impsamplesname$}
    \SetKwData{sampleirrelimps}{irrel\_imp}
    \SetKwData{xs}{$\datawithperm$}
    \SetKwData{s}{s}
    \SetKwData{X}{X}
    \SetKwData{y}{y}
    \SetKwData{NumSamples}{$N_{\text{Samples}}$}
    \SetKwData{ScoreBounds}{LossBounds}
    \SetKwData{ShadowBounds}{IrrelBounds}
    
    \SetKwFunction{Model}{Model}
    \SetKwFunction{score}{loss}
    \SetKwFunction{importance}{importance}
    \SetKwFunction{addirrel}{generatePermFeature}
    \SetKwFunction{createStatistic}{t-statistic}
    
    \SetKwInOut{Input}{Input}
    \KwData{\X, \y}
    \Input{\Model,  \NumSamples}
    
    \scores $ \gets $ \O\ (empty set)\;
    \allsampleirrelimps $ \gets $ \O \;
    
    s $\leftarrow 0$\;
    \While{$s < $\ \NumSamples; s++}{
        \s $\gets$ \addirrel{\X}\;
        \xs $\gets \X \cup \s$\;
        \;
        \tcp{Fit model}
        ext-model $\gets$ \Model{Xs, y}\;
        \;
        \tcp{loss samples \autoref{sub:loss_comparison}}
        \samplescore $\gets$ \score{ext-model, \xs, \y}\;
        \tcp{importance samples \autoref{sub:importance_statistics}}
        \sampleirrelimps $\gets$ \importance(ext-model, s)\;
        \;
        \scores $\gets$ \scores $\cup$ \samplescore\;
        \allsampleirrelimps $\gets$ \allsampleirrelimps $\cup$ \sampleirrelimps \;
    }
    
    \ScoreBounds $\gets$ \createStatistic{\scores}\;
    \ShadowBounds $\gets$ \createStatistic{\allsampleirrelimps}\;

    \KwOut{\ScoreBounds, \ShadowBounds}
    \caption{Estimating loss and irrelevant importance distribution}
    \label{alg:stat}
\end{algorithm}

\begin{algorithm}
    \DontPrintSemicolon
    \SetKwData{X}{X}
    \SetKwData{V}{\ownset{V}}
    \SetKwData{I}{\ownset{I}}
    \SetKwData{y}{y}
    \SetKwData{SRset}{\SR}
    \SetKwData{WRset}{\WR}
    \SetKwData{ARset}{\AR}
    \SetKwData{MRset}{\MR}
    \SetKwData{initscore}{$\predi$}
    \SetKwData{currloss}{$loss_j$}
    \SetKwData{redmodel}{reduced\_model}
    
    \SetKwFunction{Boruta}{Boruta}
    \SetKwFunction{RF}{RF}
    \SetKwFunction{impselection}{ImpSelection}
    \SetKwFunction{score}{loss}
    \SetKwFunction{scoreBounds}{lossBounds}
    \SetKwFunction{fit}{fit}
    \SetKwFunction{select}{select}
    \KwData{\X,\y}
    \SetKwInOut{Input}{Input}
    \Input{\RF}
    \SRset $\gets$ \O\;
    \WRset $\gets$ \O\;

    \ARset $\gets \Boruta{\RF, X, y}$\;
    \MRset $\gets \impselection {\RF, X, y}$ (\autoref{eq:minimalset})\;
    \;
    \tcp{Reduce dataset \X to relevant features}
    \V $\gets$ \select{\X, \ARset}\;
    \;
    \tcp{loss bounds using \autoref{alg:stat}}
    \initscore $\gets$ \scoreBounds{\RF, \V, \y}\;
    \BlankLine
    \tcp{iterate over subset M}
    \For{feature $\feat$ in \MRset}{
    
        \tcp{Remove current feature}
        \I $\gets$ \V without feature \feat\;
        
        \tcp{find best model without \feat}
        \redmodel $\gets$ \fit{\RF, \I, \y}\;
        
        \tcp{compute score}
        \currloss $\gets$ \score{\redmodel, \I, \y}\;
        
        \tcp{add current feature to \SRset if significantly worse}
        \If{\currloss not in \initscore}{
            \SRset $\gets \SRset \cup \{\feat\}$\;
        }
    }
    \tcp{decomposition(\autoref{eq:decomposition})}
    \WRset $\gets \ARset \setminus \SRset$\;
    \KwOut{\SRset, \WRset}
    \caption{Iterative Decomposition Algorithm}
    \label{alg:main}
\end{algorithm}

\subsection{Feature Importance Relevance Determination}
\label{sub:importance_statistics}

As mentioned earlier in \autoref{sec:rfs} we do not use existing minimal feature selection methods to decide which features are part of \MR.
Instead, we propose to use feature importance values from the Random forest model to efficiently decide which features are relevant
In contrast to Boruta, we are interested in a sparse solution of the feature set without redundant features, which represents \MR.

Deep learning models with many hidden layers can be opaque in their attribution of the input features in relation to the output layer.
In the case of random forest there exist several measures of feature importance.
They commonly express a feature's importance by averaging an information measure over all splits in the decision forest, which were part of the ensemble.
Examples are the average information gain of the objective function or the number of correct classifications with and without the feature.
In the following lets consider the information gain measure as $\featimpfunc{\data}{\optfunc}\in \mathbb{R}^\numfeatures$.
$\featimpfunc{\data}{\optfunc}_\feat$ is then the improvement of the splitting criterion averaged over all trees and all splits where feature $\feat$ is used as the split feature.

If an importance measure would correlate to the relevance of the input feature, we could use these measures in deciding which features are relevant.

For example, we expect correlated or identical features to exhibit the same feature importance but in practice some features are implicitly preferred by small differences in information or pure stochastic reasons.
We demonstrate this in \autoref{fig:imp_dists} (b) where some features have much bigger importance where others only have small or zero importance.
This is an example of correlation bias \autocite{tolosiClassificationCorrelatedFeatures2011a} which is a common problem for many importance measures in general.

One important parameter in fitting a random forest is the \emph{feature fraction} which denotes how many features are included for each bootstrap tree.
When this fraction is high ($\approx 100\%$) most features are included.
To circumvent correlation bias, we use a feature fraction of only $10\%$ in our experiments which yields a more even distribution of importances and is close to the recommended optimum of $\sqrt{\numfeatures}$ \autocite{diaz-uriarteGeneSelectionClassification2006} as is demonstrated in \autoref{fig:imp_dists} (a).

Similar to linear models, using a sparse regularization we could force redundant features to exhibit low importance in the model.
Lasso uses L1 regularization which leads to many zero entries in the model weights and features with non-zero weights are considered part of the feature set.
By parameterizing the RF to have a high \emph{feature fraction} we force a similar sparsity as can be seen in \autoref{fig:imp_dists}.
But apparent in the figure is that even irrelevant features do not have an importance value equal to zero which makes a simple threshold at zero is impossible.
Thus, important for this approach is a well-defined threshold to decide which value is considered irrelevant.
A simple measure like the mean of all importance values (blue horizontal line) does not work in general.
We therefore propose to replicate the statistical comparison with a null distribution from \autoref{sub:loss_comparison} in the context of importance values.

\subsubsection{Minimal Feature Set} %
\label{ssub:minimal_feature_set}

In \autocite{pfannschmidtFRIFeatureRelevanceIntervals2019} the authors proposed a statistics based approach to estimate a dynamic threshold for linear models.
They use randomly permuted real features to simulate irrelevant variables and observe their weights in the model over many samples.
We replicate this and extend the statistics from \autoref{sub:loss_comparison} to include the importance values of randomly permuted features.

Because we create samples $\datawithperm$ identical to the samples in $\probedist$ for the loss distribution already, we can efficiently combine both samples and model fitting at the same time.
For the distribution in \autoref{sub:loss_comparison} we focus on the score, whereby here we focus on the feature importance values\ie we use the same models to create the distributions.

The samples are then defined as
\[
    \shadowfeatprobedist{\optfunc,\data}:=\{\featimpfunc{ \perm{\feat}}{\optfunc}\ |\; \feat \sim U(\numfeatures) \}^{\numscoresamples}
\]
where $\permname$ is the permutation function and $\perm{\feat}$ is the shadow feature.
We therefore sample the importance of the shadow feature in the model $\optfunc$.
The bounds are then defined as
\[
    \shadowPI{\cdot}.
\]

To produce the minimal-optimal feature set \MR we fit a simple random forest with a high allowed feature fraction.
This leads to the behaviour seen in \autoref{fig:imp_dists} (b) where only a subset of correlated features shows significant importance.
We then compare each feature's importance value with $\shadowfeatpredi$ which represents the distribution of importance values we consider as irrelevant.
The minimal-optimal set is then given as 
\begin{equation}\label{eq:minimalset}
    \MR :=\{ \featimpfunc{\data}{\optfunc}_\feat > \shadowfeatpredi \; |\; \forall \feat \in \allfeatures \}.
\end{equation}

\begin{figure}
    \begin{small}
        \begin{center}
            \includegraphics[width=0.95\linewidth]{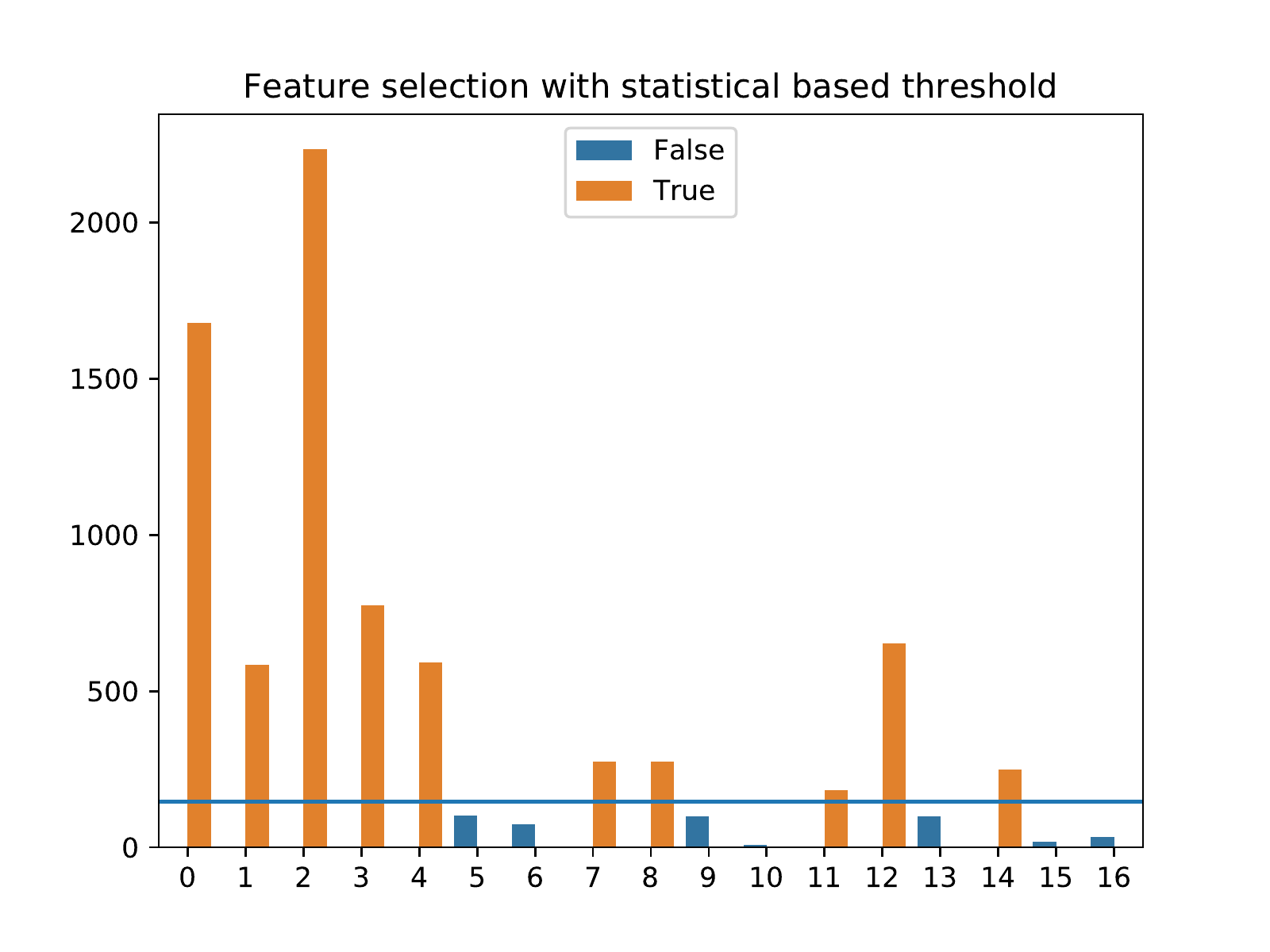}
        \end{center}
        \caption{Importance values per feature (bars) of random forest model and upper feature selection threshold of $\shadowfeatpredi$.
            The lower threshold is < 0 and excluded in this figure. 
            Colors denote the membership of features to \MR. }
        \label{fig:featselection_with_stats}
    \end{small}
\end{figure}

\autoref{fig:featselection_with_stats} shows the upper bound of the interval in use with random forest model on toy data.

\section{Results}
\label{sec:results}

\subsection{Implementation} %
\label{ssub:Implementation}
For the experiments we implemented the algorithms in \autoref{sec:methods} in Python utilizing several existing libraries.
The methods such as Boruta or our importance selection method are wrappers which require the definition of an inner model.
Because we are fitting this model many times, we opted for a very efficient random forest implementation, which is provided by the \emph{LightGBM} library \autocite{keLightGBMHighlyEfficient2017}.

The Boruta \autocite{kursaFeatureSelectionBoruta2010} method is implemented in the \emph{boruta\_py} library\footnote{\url{https://github.com/scikit-learn-contrib/boruta_py}}.
Other utility functions are used from \emph{scikit-learn} \autocite{Pedregosa2011ScikitLearn}.
The complete implementation (nicknamed \emph{"Squamish"}) and source code is publicly available\footnote{\url{https://github.com/lpfann/squamish}}.

\subsection{Evaluation} %
\label{sub:Evaluation}
In this section we try to characterize the behaviour of our proposed method.
All scripts and data generation methods are publicly available and results can be reproduced\footnote{instructions available at \url{https://github.com/lpfann/squamish_experiments}}.

\subsubsection{Selection Methods} %
\label{ssub:eval_methods}
To show the characteristics of our methods we ran several benchmarks against established feature selection methods.
Here we specifically focus on the performance of the all-relevant feature selection in a setting where redundant features are present.

We compare against the following approaches
\begin{itemize}
    \item \emph{ElasticNet}: A linear model using the \emph{ElasticNet} method which combines both $L_2$ and $L_1$ regularization. The combination can be weighted linearly and allows for more sensitivity of redundant features. In our experiments we fit this parameter through grid search in combination with cross validation.
          Features are selected according to recursive feature elimination which is guided by a cross-validated model performance \autocite{elasticnet}.
    \item \emph{RF}: A random forest model (\emph{LightGBM}) with recursive feature elimination as the selection method where number of features is also decided by cross-validation. 
    \item \emph{FRI}: The feature relevance interval method \autocite{pfannschmidtLpfannFriFeature2020,pfannschmidtFRIFeatureRelevanceIntervals2019} as the only representative with distinction between strong and weak relevance.
    \item \emph{SQ}: The sequential feature class decomposition method presented in \autoref{alg:main} and implemented as described in \autoref{ssub:Implementation}.
                        Parameters for the statistical test where $\numscoresamples=50$ and $p=10^{-6}$
\end{itemize}
Hyper-parameters for all methods where decided by cross-validation.
The tree models used the default parameters from \emph{LightGBM} except: 
\begin{description}
    \item \code{num\_leaves}: 32
    \item \code{max\_depth}: 5
    \item \code{boosting\_type}: rf
    \item \code{bagging\_fraction}: 0.632
    \item \code{bagging\_freq}: 1
\end{description}
The \code{feature\_fraction} parameter was left to default ($1$) for the \emph{RF} method. 
In \emph{SQ} for the loss comparison it was set to $0.8$. \emph{SQ} set Borutas' tree model to \code{feature\_fraction}$=0.1$.

\subsubsection{Stability of Feature Importance Values} %
\label{ssub:eval_stabil_imp_vals}

The proposed method is utilizing random forest importance values to decide about the relevance of features.
In the following we perform a short analysis of the variance of these importance values over multiple model fits on the same dataset.
We employ a random forest model as described in the section before with two different parameter choices.
The first choice is a low $\text{feature\_fraction}=0.1$ and the second the maximum $\text{feature\_fraction}=1$ such that all features are allowed in each tree generation.

We generate a simple linear classification dataset with 17 features and 300 samples.
Features 0-4 are considered strongly relevant, features 5-14 are weakly relevant with in-between correlations, and features 15-16 are irrelevant.

\begin{figure}
    \begin{subfigure}{\linewidth}
        \includegraphics[width=\linewidth]{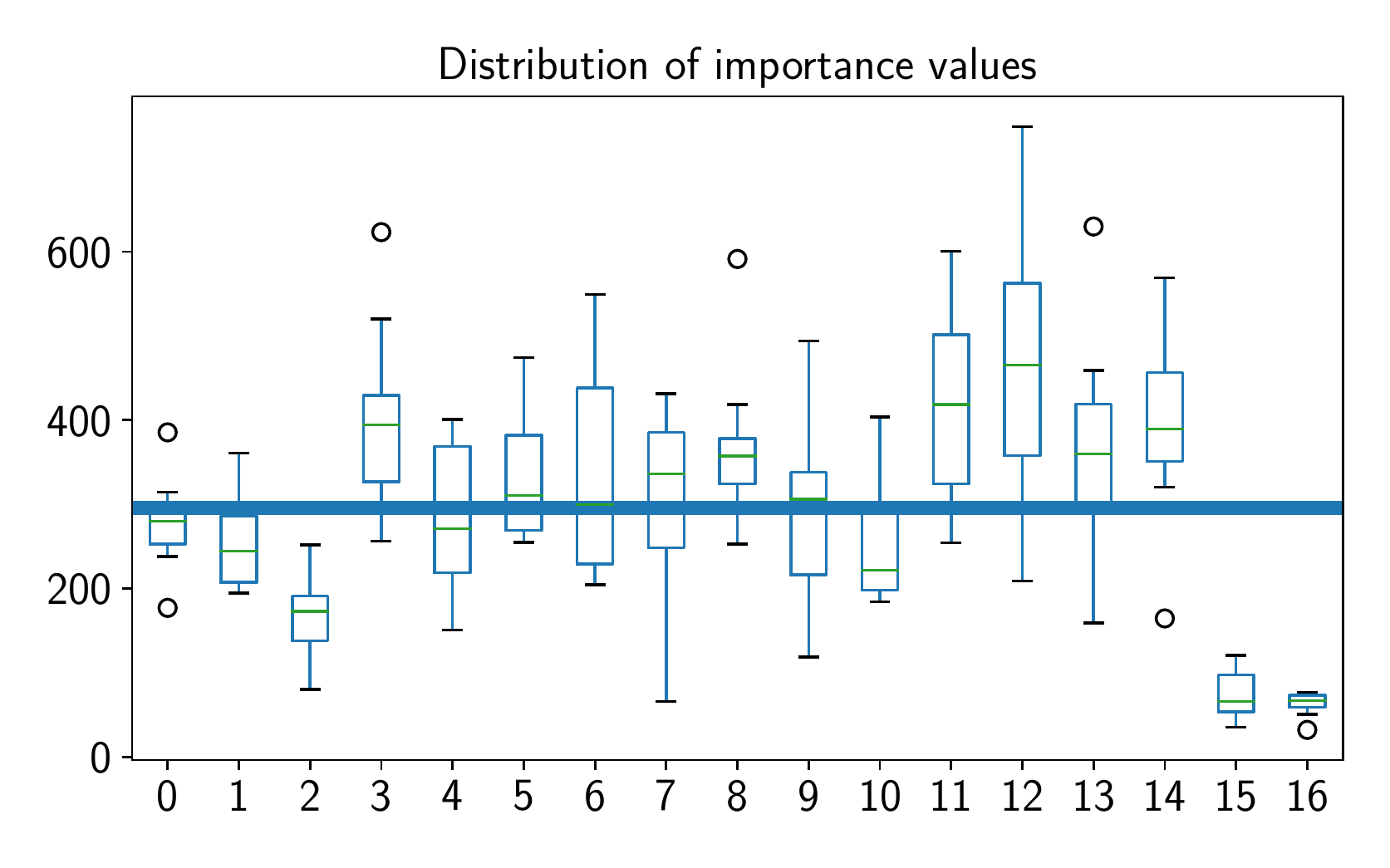}
        \caption{0.1 feature fraction}
    \end{subfigure}\par\medskip
    \begin{subfigure}{\linewidth}
        \includegraphics[width=\linewidth]{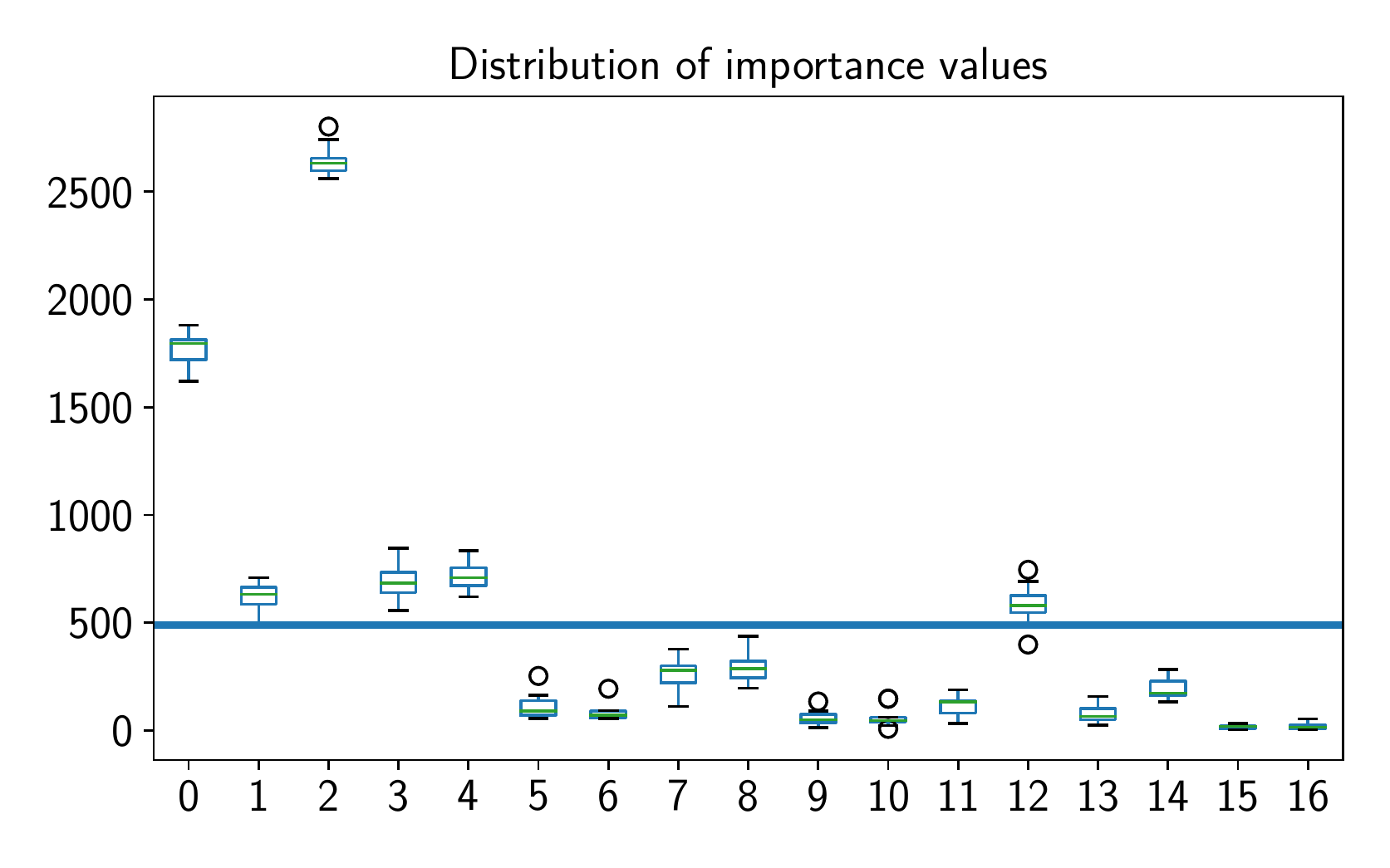}
        \caption{1.0 feature fraction}
    \end{subfigure}\par\medskip
    \caption{Distribution of feature gain importance values of random forest classifier over multiple bootstrap iterations on toy example where features 0-14 are correlated and 15-16 are irrelevant. Mean of all importance values is given as blue horizontal line. Subplots (a) and (b) represent different fraction of features allowed in tree construction.}
    \label{fig:imp_dists}
\end{figure}

The test consisted of fitting the model on the dataset 10 times in a row and record the feature importance gain measure as defined in \autoref{sub:importance_statistics}.
The resulting distributions are visualized in \autoref{fig:imp_dists} for each parameter choice.
One can see, that even without any variation of the data, the values show high variance in both settings and do not correlate to the real mutual information with the target variable.
Furthermore, the choice of a high feature fraction leads to some variables overshadowing the importance of others such as seen in feature 12 which contains the same information as all other weakly relevant in this case.
A lower fraction leads to a more evenly distributed importance signature.  

\subsubsection{Parameterization for Feature Selection} %
\label{ssub:eval_featsel_with_rf}
Based on the evaluation in \autoref{ssub:eval_stabil_imp_vals} we also perform further analysis on the consequences of the parameter for feature selection.
We extend the experiment with a feature selection step.
Compared is recursive feature elimination guided by cross validation with Boruta.

\begin{figure*}
    
    \begin{subfigure}{\columnwidth}
        \includegraphics[width=\columnwidth]{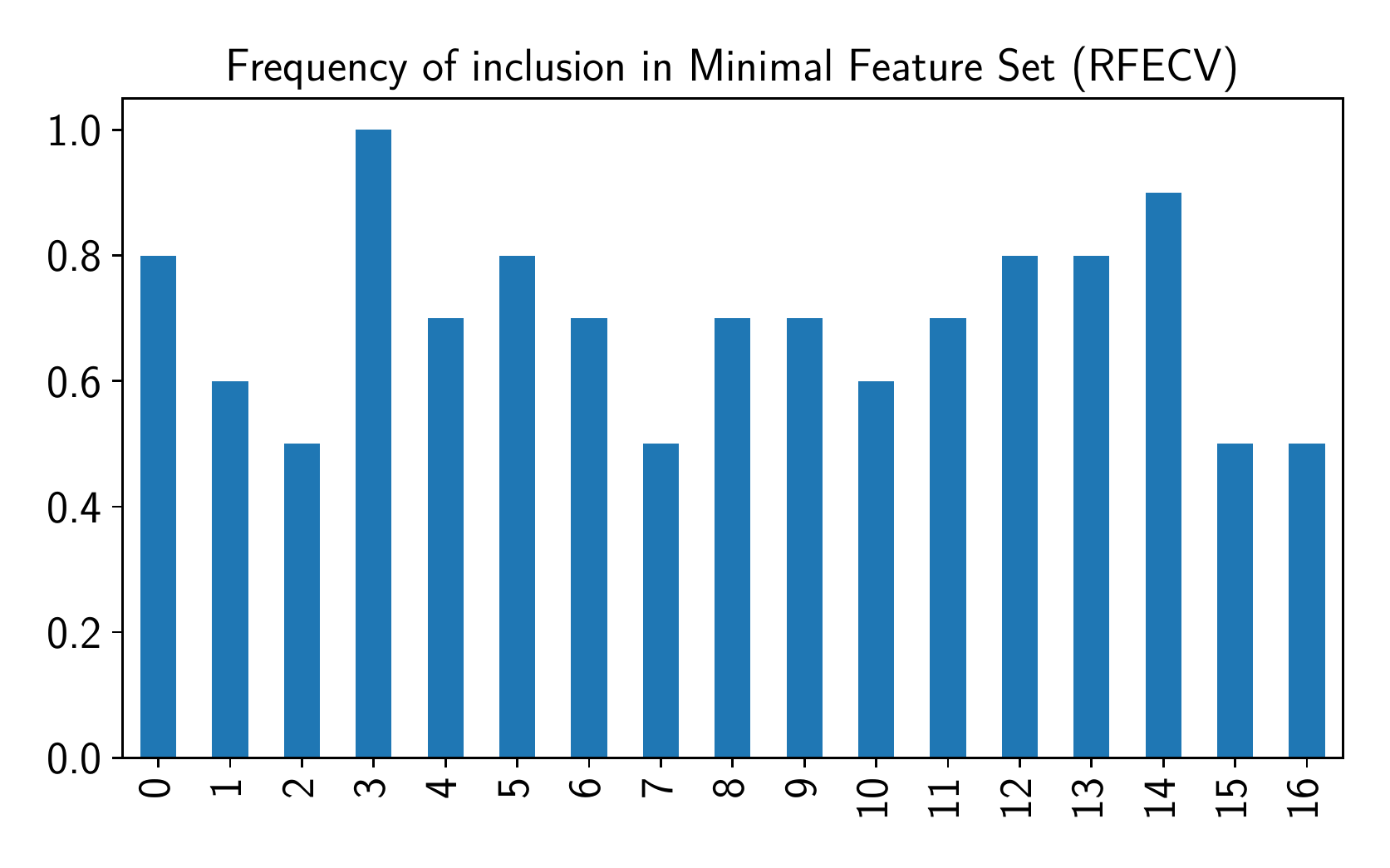}
        \caption{RFECV, $\text{feature\_fraction}=0.1$}
    \end{subfigure}
    \begin{subfigure}{\columnwidth}
        \includegraphics[width=\columnwidth]{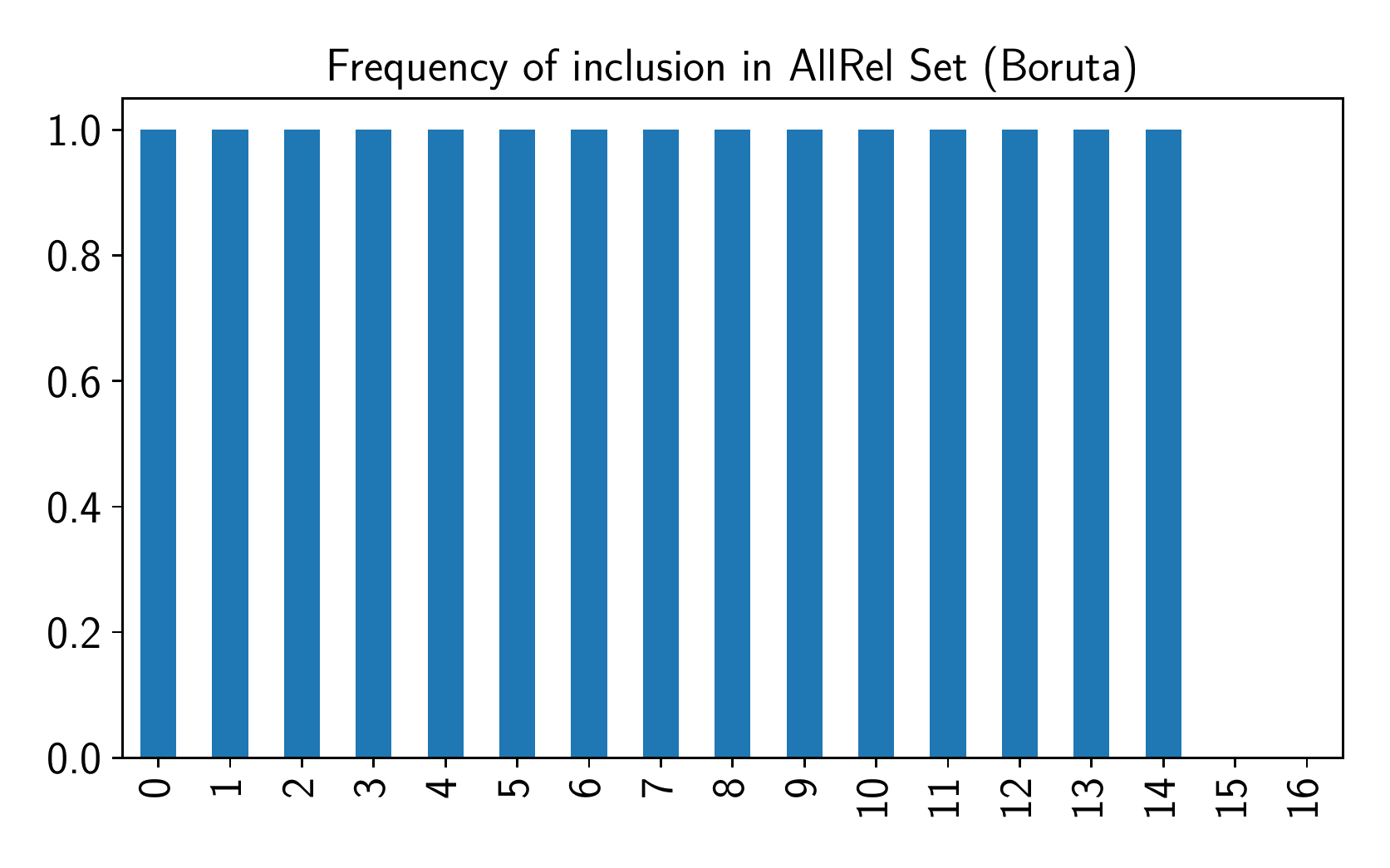}
        \caption{Boruta, $\text{feature\_fraction}=0.1$}
    \end{subfigure}
    
    \begin{subfigure}{\columnwidth}
        \includegraphics[width=\columnwidth]{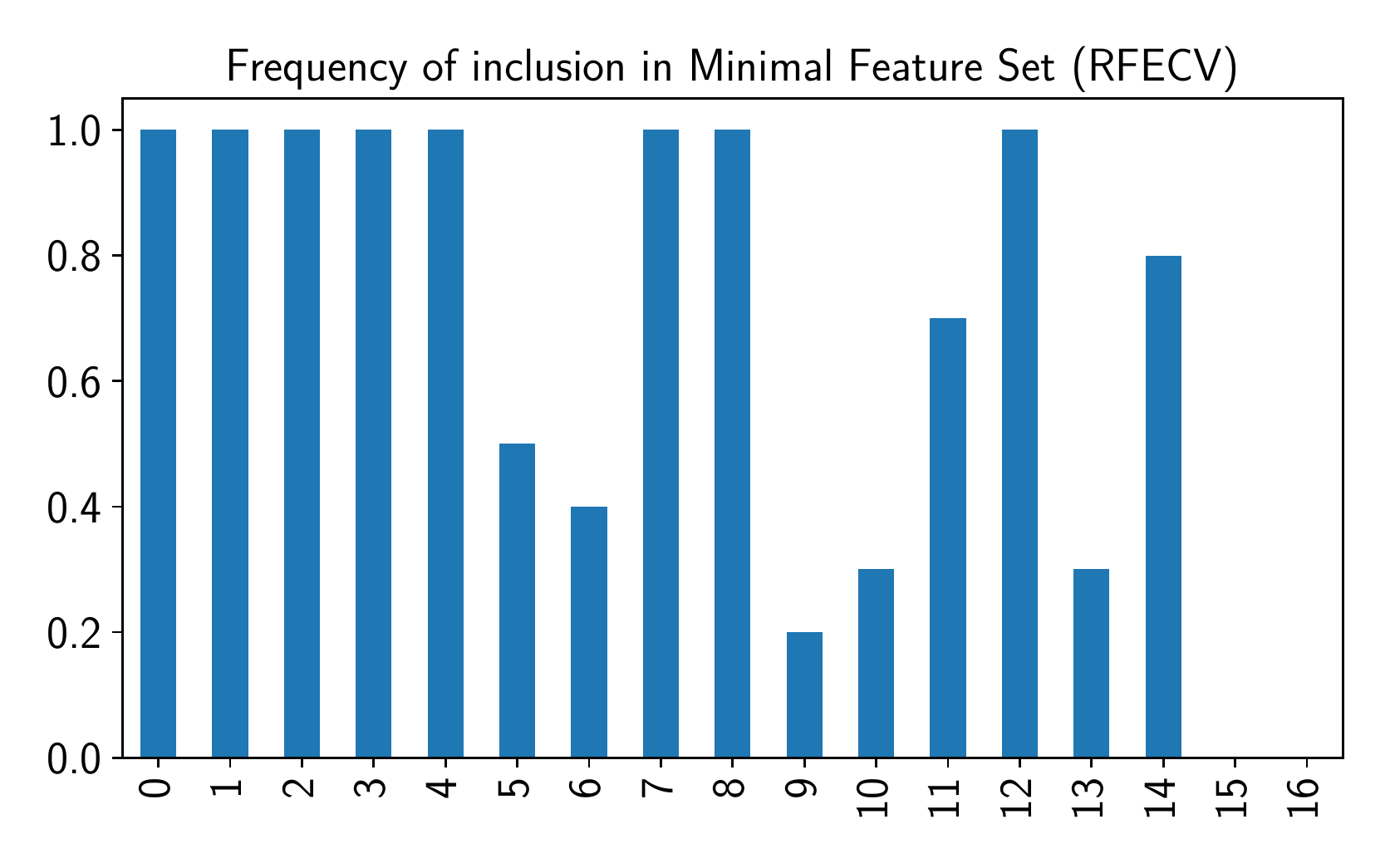}
        \caption{RFECV, $\text{feature\_fraction}=1$}
    \end{subfigure}
    \begin{subfigure}{\columnwidth}
        \includegraphics[width=\columnwidth]{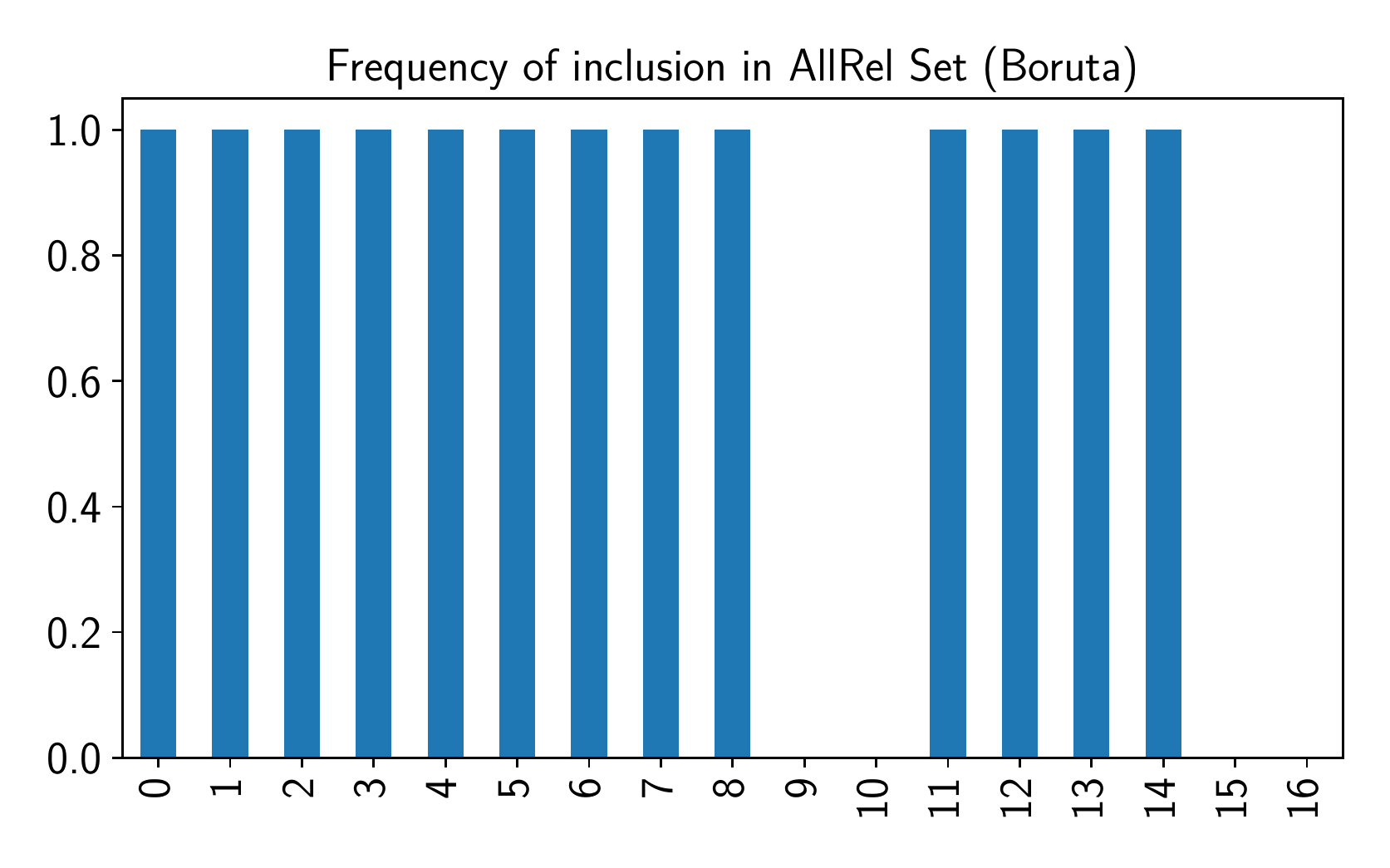}
        \caption{Boruta, $\text{feature\_fraction}=1$}
    \end{subfigure}
    \caption{Frequency of feature selection for dataset with 5 strongly relevant features (0-4), 10 weakly relevant features (5-14) and 2 irrelevant features (15-16) as described in \autoref{ssub:eval_stabil_imp_vals}. Vertical bars represent probability that each feature was included in the selected feature set for RFECV and Boruta. The random forest model used differen setting for $\text{feature\_fraction}$.}
    \label{fig:imp_frequency_featurefraction}

\end{figure*}

We record the number of times each feature was selected in the feature set.
This results in the frequency of selection or the probability that a feature is selected.
    \label{fig:imp_frequency_featurefraction}
The frequencies are given in \autoref{fig:imp_frequency_featurefraction} for both models and both parameter settings for $\text{feature\_fraction}$.
We can see that a lower feature fraction is beneficial in the case of all-relevant feature selection where the Boruta model recognizes all relevant features 0-14 without selecting random features.
The RFE procedure on the other hand suffers in this case and looses precision by selecting irrelevant features $50\%$ of the time.
It performs better with a high feature fraction parameter and selects strongly relevant features (0-4) consistently but shows higher variance in the case of weakly relevant features.

\subsubsection{General Feature Selection Accuracy} %
\label{ssub:eval_general_featsel_acc}
The most relevant metric for feature selection methods is the accuracy of selected features.
When the ground truth is known, we can explicitly evaluate the validity of the selected features.
We focus on the all-relevant feature selection problem where we use the following measures to evaluate the match of the detected feature set and the known ground truth of all relevant features:
precision and recall.
Recall is defined by \code{TP} / (\code{TP}+\code{FN}) with \code{TP} = number of true positives and \code{FN} = number of false negatives.
It denotes how many of the relevant features were selected which is crucial when looking for the all relevant feature set.
Precision is defined by \code{TP} / (\code{TP}+\code{FP}) with \code{FP} = number of false positives and describes the frequency of false positives part of the feature set.
One can use the \fmes measure as the harmonic mean of precision and recall:
\[
{\displaystyle \fmes =2\cdot {\frac {\mathrm {precision} \cdot \mathrm {recall} }{\mathrm {precision} +\mathrm {recall} }}}
\]

In this evaluation we compare the methods from \autoref{ssub:eval_methods} by highlighting the \fmes measure.

First we compare all methods on a linear classification dataset to allow a fair comparison with the linear models.
To generate a multidimensional classification problem, we use a randomly generated prototype vector which defines a hyperplane.
The defining features of this plane are strongly relevant.
Sample points are generated in this feature space and the class is determined by the side of the hyperplane the points lie on.
Weakly relevant features are constructed by replacing a feature of the original feature space with its linear combination.
The elements of this combination are highly correlated and produce a set of redundant features.
By removing the original feature and replacing it with those elements we achieve weak relevance by definition.
Irrelevant features are sampled from a standard normal distribution.

We generate 8 datasets with different feature set compositions as given by \autoref{tab:toyparams}.
For example, Set 1 consists of 150 samples (n), 6 strongly relevant features (strong), no weak relevant features (weak) and 6 irrelevant random features (irr).

All models given in \autoref{ssub:eval_methods} are repeatedly fit on bootstraps of these datasets, resulting in 10 results per dataset per model which are averaged in the following.
The prediction accuracy on the datasets is listed in \autoref{tab:trainacc} with sufficient accuracy for all models.

The results of the \fmes measure evaluation are given in \autoref{tab:toyscoref1}.
Additionally, we recorded the runtime of all methods while performing feature selection which is given in \autoref{tab:runtime}.

Most evident is the perfect score of FRI in this setting while being the slowest method.
SQ follows second and performs very good feature selection in all cases while being the second fastest.
The RFE scheme using random forests (RF) performed worst, not selecting weakly relevant features such as in Set 6 and 7.
The ElasticNet also not as sensitive in this experiment, but showing the fastest runtime given its simplicity.

\begin{table}
    \begin{small}
        \caption{Parameters of synthetically generated datasets for a \emph{linear separable} classification problem as described in \autoref{ssub:eval_general_featsel_acc}.
                Columns denote number of features with corresponding characteristics:
                \emph{n} (number of samples), \emph{strong} (number of strongly relevant features), \emph{weak} (number of weakly relevant features),
                \emph{irr} (number of irrelevant features).
                }
        \label{tab:toyparams}
        \begin{center}
            \begin{tabular}{lrrrr}
\toprule
{} &     n &  strong &  weak &  irr \\
\textbf{Set  } &       &         &       &      \\
\midrule
\textbf{Set 1} &   150 &       6 &     0 &    6 \\
\textbf{Set 2} &   150 &       0 &     6 &    6 \\
\textbf{Set 3} &   150 &       3 &     4 &    3 \\
\textbf{Set 4} &   256 &       6 &     6 &    6 \\
\textbf{Set 5} &   512 &       1 &     2 &   11 \\
\textbf{Set 6} &   200 &       1 &    20 &    0 \\
\textbf{Set 7} &   200 &       1 &    20 &   20 \\
\textbf{Set 8} &  2000 &      10 &    10 &   50 \\
\bottomrule
\end{tabular}

        \end{center}
    \end{small}
\end{table}

\begin{table}
    \begin{small}
        \caption{Training accuracy of models on \emph{linear separable} classification data generated according to \autoref{tab:toyparams}.}
        \label{tab:trainacc}
        \begin{center}
            \begin{tabular}{lrrrr}
\toprule
{} &  ElasticNet &  FRI &   RF &   SQ \\
\midrule
\textbf{Set 1} &        0.97 & 0.99 & 0.83 & 1.00 \\
\textbf{Set 2} &        0.99 & 0.99 & 1.00 & 1.00 \\
\textbf{Set 3} &        0.98 & 0.99 & 0.86 & 1.00 \\
\textbf{Set 4} &        0.98 & 0.99 & 0.87 & 1.00 \\
\textbf{Set 5} &        0.98 & 1.00 & 0.95 & 1.00 \\
\textbf{Set 6} &        0.97 & 1.00 & 0.89 & 0.99 \\
\textbf{Set 7} &        0.98 & 0.99 & 0.90 & 1.00 \\
\textbf{Set 8} &        0.98 & 1.00 & 0.91 & 1.00 \\
\bottomrule
\end{tabular}

        \end{center}
    \end{small}
\end{table}

\begin{table}
    \begin{small}
        \caption{Average \fmes measure on \emph{linear separable} data sets regarding feature classification.}
        \label{tab:toyscoref1}
        \begin{center}
            \begin{tabular}{lllrrrr}
\toprule
      &    & \textbf{model} &  ElasticNet &  FRI &   RF &   SQ \\
{} & \textbf{type} & \textbf{data} &             &      &      &      \\
\midrule
\multirow{8}{*}{\textbf{score}} & \multirow{8}{*}{\textbf{f1}} & \textbf{Set 1} &        0.91 & 1.00 & 0.86 & 0.98 \\
      &    & \textbf{Set 2} &        0.75 & 1.00 & 0.29 & 0.92 \\
      &    & \textbf{Set 3} &        0.83 & 1.00 & 0.67 & 0.97 \\
      &    & \textbf{Set 4} &        0.86 & 1.00 & 0.68 & 0.93 \\
      &    & \textbf{Set 5} &        0.85 & 1.00 & 0.77 & 0.99 \\
      &    & \textbf{Set 6} &        0.52 & 1.00 & 0.17 & 0.99 \\
      &    & \textbf{Set 7} &        0.38 & 1.00 & 0.17 & 0.95 \\
      &    & \textbf{Set 8} &        0.83 & 1.00 & 0.65 & 0.99 \\
\bottomrule
\end{tabular}

        \end{center}
    \end{small}
\end{table}

\begin{table}
    \begin{small}
        \caption{Runtime in seconds (rounded) for experiment described in \autoref{ssub:eval_general_featsel_acc}.}
        \label{tab:runtime}
        \begin{center}
            \begin{tabular}{llrrrr}
\toprule
        & \textbf{model} &  ElasticNet &  FRI &   RF &  SQ \\
{} & \textbf{data} &             &      &      &     \\
\midrule
\multirow{8}{*}{\textbf{runtime}} & \textbf{Set 1} &           0 &    2 &    0 &   1 \\
        & \textbf{Set 2} &           0 &    2 &    0 &   1 \\
        & \textbf{Set 3} &           0 &    2 &    0 &   1 \\
        & \textbf{Set 4} &           0 &    3 &    1 &   3 \\
        & \textbf{Set 5} &           0 &    5 &    2 &   6 \\
        & \textbf{Set 6} &           0 &    4 &    1 &   2 \\
        & \textbf{Set 7} &           0 &    6 &    3 &   3 \\
        & \textbf{Set 8} &           2 &  201 &  138 &  80 \\
\bottomrule
\end{tabular}

        \end{center}
    \end{small}
\end{table}

\subsubsection{Non-linear problems} %
\label{ssub:Non-linear problems}
While many problems can be tackled using linear models, many relations are non-linear in nature.
In this experiment we generate data which can not be separated with a linear hyperplane.
Our assumption is, that the linear models ElasticNet and FRI should not perform good in this case.

We utilize the classification data generation function from scikit-learn\footnote{\url{https://scikit-learn.org/stable/modules/generated/sklearn.datasets.make_classification.html}}
\ to create binary classification data with multiple opposing cluster of samples (parameter \code{n\_clusters\_per\_class} $ =2$).
We then process the informative features to produce weakly relevant (redundant) features and additional irrelevant features\footnote{Generation function available at \url{https://github.com/lpfann/arfs_gen}}.

Again, we generate sets with different feature configurations given in \autoref{tab:toy_NL_preset}.
We fit all models on 20 newly generated sets and compute the average metric values.

The combined metrics are given in \autoref{tab:stats_NL} per dataset and more concise in \autoref{tab:stats_NL_mean} averaged over all sets.
Both linear models show low training accuracy at ~70\% which hints that the linear models can not replicate the non-linear relation.
The random forest based models (SQ, RF) fare better with accuracies $\geq 83\%$.
While not exceptional, it highlights the difficulty of this toy classification problem.

\begin{table}
    \begin{small}
        \caption{Parameters of generated datasets for \emph{non-linearly} separable classification data. 
        Numeric difference between \emph{n\_features} and strong (\emph{n\_strel}) and weak (\emph{n\_redundant}) relevant features is filled with irrelevant features\ie NL~1 contains 10 irrelevant features.
        }
        \label{tab:toy_NL_preset}
        \begin{center}
            \begin{tabular}{lrrrr}
\toprule
\textbf{Set} &  NL 1 &  NL 2 &  NL 3 &  NL 4 \\
\midrule
\textbf{n\_features } &    20 &    20 &    50 &    80 \\
\textbf{n\_strel    } &    10 &     4 &    10 &    10 \\
\textbf{n\_redundant} &     0 &    10 &    10 &    10 \\
\bottomrule
\end{tabular}

        \end{center}
    \end{small}
\end{table}

\begin{table}
    \begin{small}
        \caption{Statistics of benchmark with \emph{non-linearly} separable classification data as generated according to \autoref{tab:toy_NL_preset}.
        \emph{Precision}, \emph{recall} and \emph{f1} quantify the feature selection performance, whereby \emph{accuracy} denotes the training accuracy (quality of model fit).}
        \label{tab:stats_NL}
        \begin{center}
            \begin{tabular}{llrrrr}
\toprule
         & \textbf{model} &  ElasticNet &  FRI &   RF &   SQ \\
{} & \textbf{dataset} &             &      &      &      \\
\midrule
\multirow{4}{*}{\textbf{precision}} & \textbf{NL 1} &        0.88 & 1.00 & 0.71 & 0.97 \\
         & \textbf{NL 2} &        0.86 & 1.00 & 0.67 & 1.00 \\
         & \textbf{NL 3} &        0.41 & 1.00 & 1.00 & 0.80 \\
         & \textbf{NL 4} &        0.21 & 1.00 & 1.00 & 0.62 \\
\cline{1-6}
\multirow{4}{*}{\textbf{recall}} & \textbf{NL 1} &        0.70 & 0.53 & 1.00 & 1.00 \\
         & \textbf{NL 2} &        0.86 & 1.00 & 0.86 & 1.00 \\
         & \textbf{NL 3} &        0.55 & 0.77 & 0.55 & 0.63 \\
         & \textbf{NL 4} &        0.50 & 0.89 & 0.45 & 1.00 \\
\cline{1-6}
\multirow{4}{*}{\textbf{f1}} & \textbf{NL 1} &        0.78 & 0.69 & 0.83 & 0.98 \\
         & \textbf{NL 2} &        0.86 & 1.00 & 0.75 & 1.00 \\
         & \textbf{NL 3} &        0.47 & 0.87 & 0.71 & 0.70 \\
         & \textbf{NL 4} &        0.30 & 0.94 & 0.62 & 0.77 \\
\cline{1-6}
\multirow{4}{*}{\textbf{accuracy}} & \textbf{NL 1} &        0.66 & 0.67 & 0.79 & 0.82 \\
         & \textbf{NL 2} &        0.70 & 0.75 & 0.81 & 0.83 \\
         & \textbf{NL 3} &        0.60 & 0.66 & 0.87 & 0.89 \\
         & \textbf{NL 4} &        0.73 & 0.74 & 0.86 & 0.90 \\
\bottomrule
\end{tabular}

        \end{center}
    \end{small}
\end{table}
\begin{table}
    \begin{small}
        \caption{Mean over all datasets of benchmark with non-linearly separable classification data as in \autoref{tab:stats_NL}.}
        \label{tab:stats_NL_mean}
        \begin{center}
            \begin{tabular}{lrrrr}
\toprule
{} &  precision &  recall &   f1 &  accuracy \\
\textbf{model     } &            &         &      &           \\
\midrule
\textbf{ElasticNet} &       0.59 &    0.65 & 0.60 &      0.67 \\
\textbf{FRI       } &       1.00 &    0.80 & 0.88 &      0.71 \\
\textbf{RF        } &       0.85 &    0.71 & 0.73 &      0.83 \\
\textbf{SQ        } &       0.85 &    0.91 & 0.86 &      0.86 \\
\bottomrule
\end{tabular}

        \end{center}
    \end{small}
\end{table}

First we analyse the general feature selection accuracy without discriminating between the relevance class subsets.
Overall the selection accuracy of all methods get worse which is apparent in \autoref{tab:stats_NL_mean}.
The ElasticNet scores last, with an average recall of $0.65$ followed by the recursive feature elimination method (RF) with $0.71$.
It is beat by FRI with recall of $0.8$ even though it can not handle non-linear separable data.
SQ has the highest recall with 91\% of relevant features recognized.
When also considering the precision, we see that FRI scores perfect precision with no false positives such that the overall result actually is better than SQ here with an \fmes of $0.88$ which is slightly better than SQ.

\subsubsection{Accuracy of Relevance Classification} %
\label{ssub:Feature Classification Accuracy}
The general feature selection accuracy evaluation in the sections before does not consider the difference between strong and weak relevance.
From all models considered before, only FRI and SQ provide can provide this distinction.
We now present the precision and recall on the subsets \SR and \WR recorded in the experiments in the previous evaluations on linearly and non-linearly separable datasets.
For that we reuse the same metrics as before and independently evaluate each subset.

\begin{table}
    \begin{small}
        \caption{Analysis of feature selection accuracy grouped by relevance class subsets on \emph{linearly separable data}.}
        \label{tab:prec_recall_arfs}
        \begin{center}
            \begin{tabular}{llrrrr}
\toprule
       &      & \multicolumn{2}{l}{Weakly} & \multicolumn{2}{l}{Strongly} \\
       &      &    FRI &   SQ &      FRI &   SQ \\
\midrule
\multirow{8}{*}{\textbf{precision}} & \textbf{Set 1} &   1.00 & 1.00 &     1.00 & 1.00 \\
       & \textbf{Set 2} &   1.00 & 0.87 &     1.00 & 1.00 \\
       & \textbf{Set 3} &   1.00 & 0.99 &     1.00 & 0.74 \\
       & \textbf{Set 4} &   1.00 & 0.99 &     1.00 & 0.86 \\
       & \textbf{Set 5} &   0.99 & 0.99 &     1.00 & 0.49 \\
       & \textbf{Set 6} &   1.00 & 1.00 &     1.00 & 0.50 \\
       & \textbf{Set 7} &   1.00 & 0.99 &     1.00 & 0.46 \\
       & \textbf{Set 8} &   0.99 & 1.00 &     1.00 & 0.90 \\
\cline{1-6}
\multirow{8}{*}{\textbf{recall}} & \textbf{Set 1} &   1.00 & 1.00 &     1.00 & 0.95 \\
       & \textbf{Set 2} &   1.00 & 1.00 &     1.00 & 1.00 \\
       & \textbf{Set 3} &   1.00 & 0.72 &     1.00 & 0.99 \\
       & \textbf{Set 4} &   1.00 & 0.68 &     1.00 & 0.99 \\
       & \textbf{Set 5} &   1.00 & 0.50 &     1.00 & 1.00 \\
       & \textbf{Set 6} &   1.00 & 0.93 &     1.00 & 1.00 \\
       & \textbf{Set 7} &   1.00 & 0.88 &     1.00 & 1.00 \\
       & \textbf{Set 8} &   1.00 & 0.85 &     1.00 & 1.00 \\
\bottomrule
\end{tabular}

        \end{center}
    \end{small}
\end{table}

\begin{table}
    \begin{small}
        \caption{Mean feature selection accuracy metrics grouped by relevance class subsets and averaged over all \emph{linearly separable} datasets. (Detailed: \autoref{tab:prec_recall_arfs})}
        \label{tab:prec_recall_arfs_mean}
        \begin{center}
            \begin{tabular}{lrrrr}
\toprule
{} & \multicolumn{2}{l}{Weakly} & \multicolumn{2}{l}{Strongly} \\
{} &    FRI &   SQ &      FRI &   SQ \\
\midrule
\textbf{precision} &   1.00 & 0.98 &     1.00 & 0.74 \\
\textbf{recall   } &   1.00 & 0.82 &     1.00 & 0.99 \\
\bottomrule
\end{tabular}

        \end{center}
    \end{small}
\end{table}

In \autoref{tab:prec_recall_arfs} we see the metrics for all linear datasets and in \autoref{tab:prec_recall_arfs_mean} their mean.
The recall for strongly relevant features is near perfect for our proposed method (SQ).
The precision is not perfect though and sometimes FP selections occur, as can be seen in Set 2 where the recall is 100\% (all relevant features where selected) but the precision is at 87\% which hints at irrelevant features being selected as well.
Additionally, in some cases such as in Set 5 SQ tends to select weakly relevant features as strongly relevant.

We also compare both methods in the much harder task from \autoref{ssub:Non-linear problems}.
The detailed results are given in \autoref{tab:feature_class_accuracy} and their average in \autoref{tab:feature_class_accuracy_mean}.
\begin{table}
    \begin{small}
        \caption{Analysis of feature selection accuracy grouped by relevance class subsets on \emph{non-linearly} separable data.}
        \label{tab:feature_class_accuracy}
        \begin{center}
            \begin{tabular}{llrrrr}
\toprule
     &    & \multicolumn{2}{l}{Weakly} & \multicolumn{2}{l}{Strongly} \\
     &    & precision & recall & precision & recall \\
\textbf{dataset} & \textbf{model} &           &        &           &        \\
\midrule
\multirow{2}{*}{\textbf{NL 1}} & \textbf{FRI} &         - &      - &      1.00 &   0.11 \\
     & \textbf{SQ} &         - &      - &      1.00 &   0.59 \\
\cline{1-6}
\multirow{2}{*}{\textbf{NL 2}} & \textbf{FRI} &      0.83 &   1.00 &      1.00 &   0.50 \\
     & \textbf{SQ} &      0.99 &   1.00 &      1.00 &   0.97 \\
\cline{1-6}
\multirow{2}{*}{\textbf{NL 3}} & \textbf{FRI} &      0.67 &   1.00 &      1.00 &   0.04 \\
     & \textbf{SQ} &      0.20 &   0.14 &      0.86 &   0.87 \\
\cline{1-6}
\multirow{2}{*}{\textbf{NL 4}} & \textbf{FRI} &      0.57 &   1.00 &      1.00 &   0.01 \\
     & \textbf{SQ} &      0.18 &   0.39 &      0.65 &   0.87 \\
\bottomrule
\end{tabular}

        \end{center}
    \end{small}
\end{table}
\begin{table}
    \begin{small}
        \caption{Mean feature selection accuracy metrics grouped by relevance class subsets and averaged over all \emph{non-linearly} separable datasets.
        (Detailed: \autoref{tab:feature_class_accuracy})}
        \label{tab:feature_class_accuracy_mean}
        \begin{center}
            \begin{tabular}{lrrrr}
\toprule
{} & \multicolumn{2}{l}{Weakly} & \multicolumn{2}{l}{Strongly} \\
{} & precision & recall & precision & recall \\
\textbf{model} &           &        &           &        \\
\midrule
\textbf{FRI  } &      0.69 &   1.00 &      1.00 &   0.16 \\
\textbf{SQ   } &      0.46 &   0.51 &      0.88 &   0.82 \\
\bottomrule
\end{tabular}

        \end{center}
    \end{small}
\end{table}

Here the results are mixed.
FRI achieves perfect recall for weakly relevant features but misses a lot of strongly relevant ones.
This is extreme in sets NL 3 and NL 4 where it has an average recall of $0.04$ and $0.01$.
Apparently, it is more inclined to classify strongly relevant features as weakly relevant because the precision of the latter is decreased.
SQ is also showing many false negative weakly relevant features while also selecting false positives which hurts its score it that setting.
On the other hand it is quite balanced in the classification of strongly relevant features and correctly selects and classifies over 80\% of them.

\section{Conclusion}
In this paper we presented a new all-relevant feature selection approach which builds upon existing methods.
We combine several ideas to decompose the all-relevant feature set into strongly and weakly relevant features.
We highlight the general selection accuracy in the linear and non-linear case which outperforms many existing approaches.
While it is not as accurate as FRI in the linear case, our method is nearly equal to FRI in the non-linear case.
In general, it performs faster and scales better to bigger datasets than FRI.
Our method also enables the classification of feature relevance classes, which is new in the case of non-linear models.
Here the results are more mixed and further analysis in correct parameterization to control false positives is needed.
Though the ratio of false positives and false negatives is always a trade off which should be considered in application.

\printbibliography
\end{document}